\documentclass{article}

\usepackage[preprint]{neurips_2025}

\usepackage[utf8]{inputenc} % allow utf-8 input
\usepackage[T1]{fontenc}    % use 8-bit T1 fonts
\usepackage{hyperref}       % hyperlinks
\usepackage{url}            % simple URL typesetting
\usepackage{booktabs}       % professional-quality tables
\usepackage{amsfonts}       % blackboard math symbols
\usepackage{nicefrac}       % compact symbols for 1/2, etc.
\usepackage{microtype}      % microtypography
\usepackage{xcolor}         % colors
\usepackage{graphicx}
\usepackage{amsmath}
\usepackage{amssymb}
\usepackage{cleveref}

\title{SkyReels-V3 Technique Report}

\author{%
SkyReels Team
}

\begin{document}
\maketitle

\begin{abstract}
  Video generation serves as a cornerstone for building world models, where multimodal contextual inference stands as the defining test of capability.
  In this end, we present SkyReels-V3, a conditional video generation model, built upon a unified multimodal in-context learning framework with diffusion Transformers. 
  SkyReels-V3 model supports three core generative paradigms within a single architecture: reference images-to-video synthesis, video-to-video extension and audio-guided video generation. (i) reference images-to-video model is designed to produce high-fidelity videos with strong subject identity preservation, temporal coherence, and narrative consistency. To enhance reference adherence and compositional stability, we design a comprehensive data processing pipeline that leverages cross-frame pairing, image editing, and semantic rewriting, effectively mitigating copy–paste artifacts. During training, an image–video hybrid strategy combined with multi-resolution joint optimization is employed to improve generalization and robustness across diverse scenarios. 
  (ii) video extension model integrates spatio-temporal consistency modeling with large-scale video understanding, enabling both seamless single-shot continuation and intelligent multi-shot switching with professional cinematographic patterns. 
  (iii) Talking avatar model supports minute-level audio-conditioned video generation by training first-and-last frame insertion patterns and reconstructing key-frame inference paradigms. On the basis of ensuring visual quality, synchronization of audio and videos has been optimized. 
  Extensive evaluations demonstrate that SkyReels-V3 achieves state-of-the-art or near state-of-the-art performance on key metrics including visual quality, instruction following, and specific aspect metrics, approaching leading closed-source systems. Github: \url{https://github.com/SkyworkAI/SkyReels-V3}.
\end{abstract}

\section{Introduction}

World models aim to capture, simulate, and forecast the dynamics of complex real-world environments, and they form a fundamental basis for deploying artificial intelligence in practical scenarios \cite{matsuo2022deep}. In between, video generation frameworks encode rich geometric, semantic, and physical knowledge through the synthesis of visual sequences, thereby enabling effective modeling and prediction of the physical world, especially for multi-modal conditions.
In recent years, diffusion-based Transformers \cite{peebles2023scalable,fei2024scaling} architecture has driven significant advances in video generation. A wide range of commercial systems—including Veo \cite{veo}, Sora \cite{sora}, Seedance \cite{gao2025seedance,chen2025seedance}, Kling \cite{Kling}, as well as open-source models such as Wanx \cite{wan2025wan}, HunyuanVideo \cite{wu2025hunyuanvideo,kong2024hunyuanvideo}, SkyReels \cite{chen2025skyreels}, and CogVideoX \cite{yang2024cogvideox}, have demonstrated strong performance across multiple dimensions. However, multimodal in-context in video generation is still under-explored. 

In this repport, we introduce SkyReels-V3, a unified multimodal condition video generation framework, designed to support a wide range of high-quality video synthesis tasks within a single model family. Built upon a multimodal in-context learning paradigm, SkyReels V3 seamlessly integrates visual reference, video, audio, and textual inputs to enable flexible and controllable video generation. The framework natively supports three core capabilities: reference images to video, video-to-video extension, and audio-guided video generation, also known as talking avatar.
At the architectural level, SkyReels-V3 incorporates large-scale diffusion Transformers while with carefully designed alignment strategies with multimodal condition and advanced spatiotemporal consistency modeling. 
Through hybrid image–video as well as multi-resolution joint optimization, the system achieves precise instruction following, high-fidelity motion generation and superior sub-domain capacity such as robust identity preservation and precise audio-visual synchronization. These design choices allow SkyReels-V3 to move beyond frame-level synthesis, enabling coherent narrative progression and cinematic-quality visual composition.

With its strong generalization ability and modular design, SkyReels-V3 can be applied to diverse real-world scenarios, including professional video production, virtual avatars, short-form content creation, live-stream commerce, and digital entertainment. Extensive evaluations demonstrate that SkyReels-V3 reaches or surpasses industry-leading performance across key metrics, making it a powerful open-source foundation for next-generation video generation research and applications.

\section{Methods and Evaluation}

The SkyReels-V3 model family supports a range of capabilities, including  multi-reference image-to-video synthesis, audio-guided video generation, and video-to-video extension. This chapter provides a detailed descriptions and performance evaluation to these features.

\subsection{Reference Images to Video}

\begin{figure*}[t]
\centering
\includegraphics[width=0.99\textwidth]{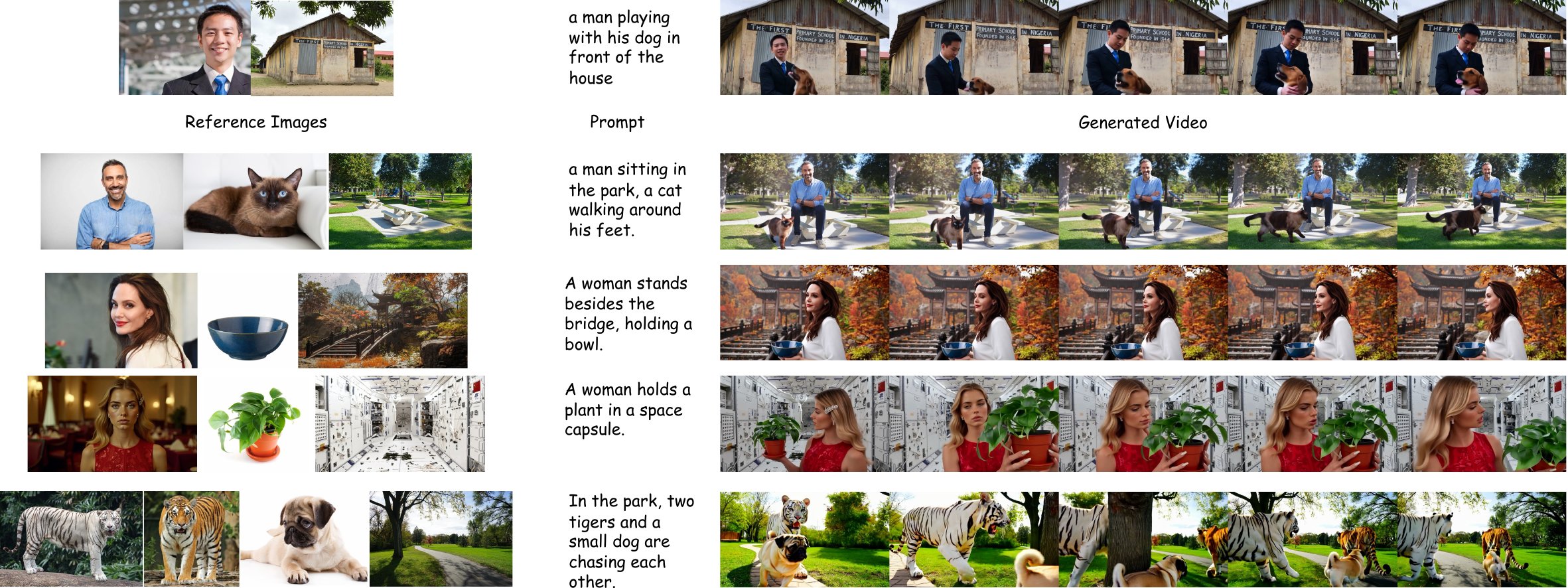}
\caption{\textbf{Reference Images to Video Results.} SkyReels-V3 can facilitate dynamic interplay between different subjects within specified contexts.}
\label{fig:mo2v1}
\end{figure*}

\begin{figure*}[t]
\centering
\includegraphics[width=0.99\textwidth]{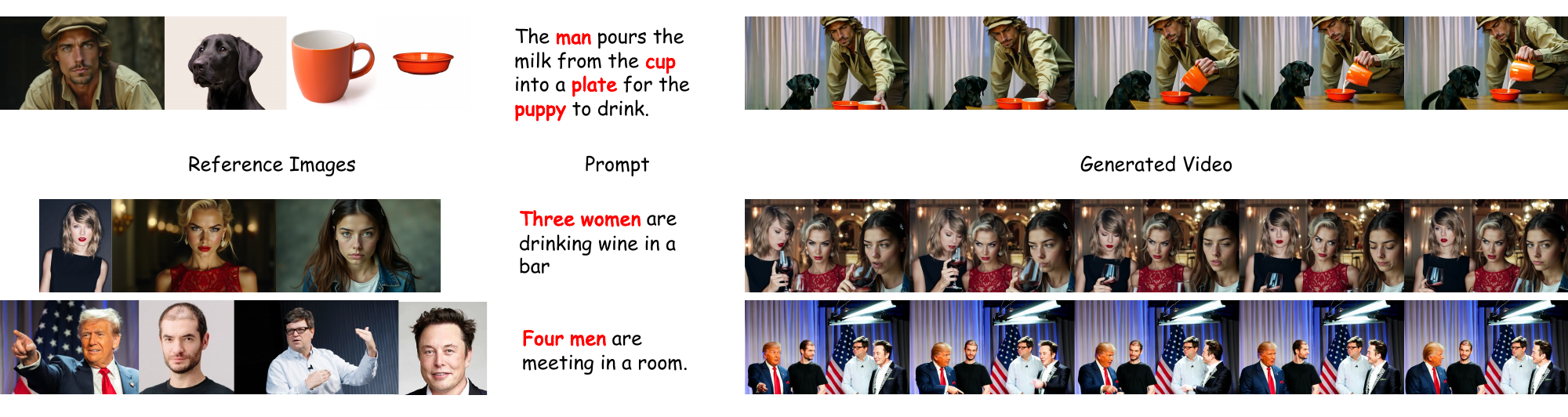}
\caption{\textbf{Reference Images to Video Results.} SkyReels-V3 can enable dynamic interactions between diverse subjects (characters/objects) within arbitrary scenes.}
\label{fig:mo2v2}
\end{figure*}

\begin{figure*}[t]
\centering
\includegraphics[width=0.99\textwidth]{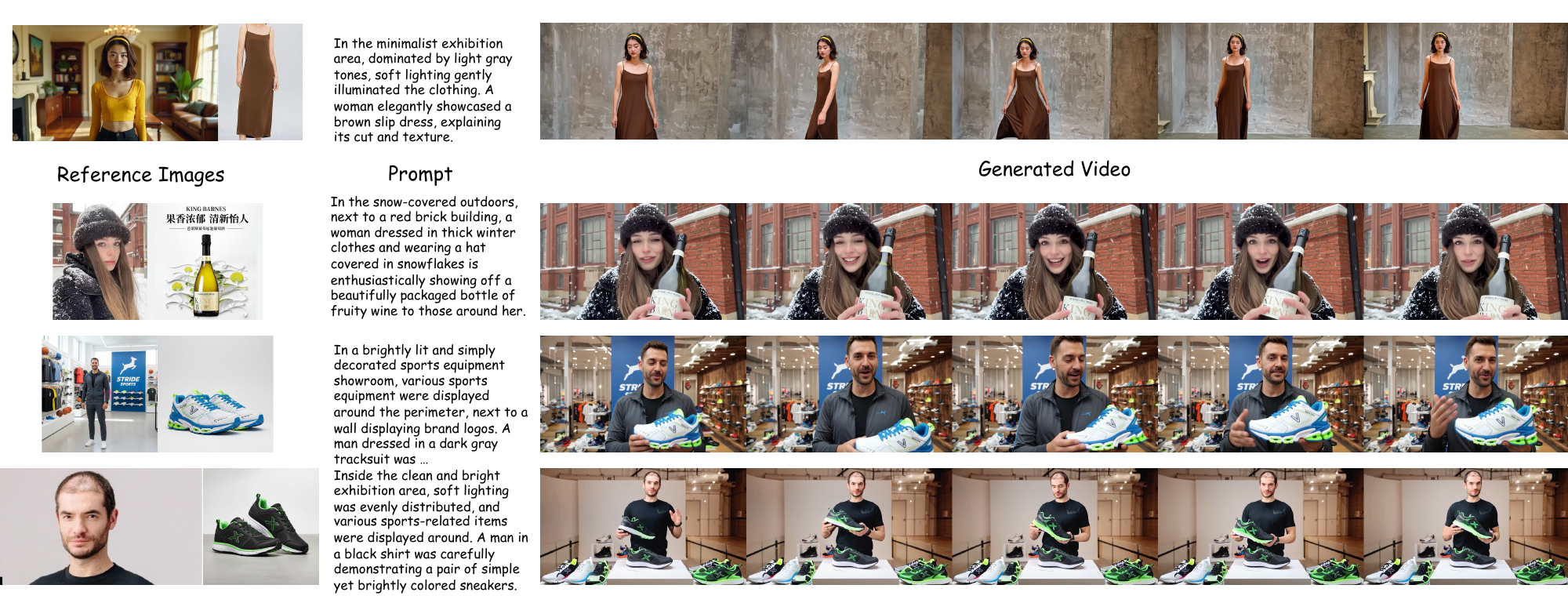}
\caption{\textbf{Reference Images to Video Results.} SkyReels-V3 can enable instant video creation for diverse live commerce hosts and settings.}
\label{fig:mo2v3}
\end{figure*}

\begin{figure*}[t]
\centering
\includegraphics[width=0.99\textwidth]{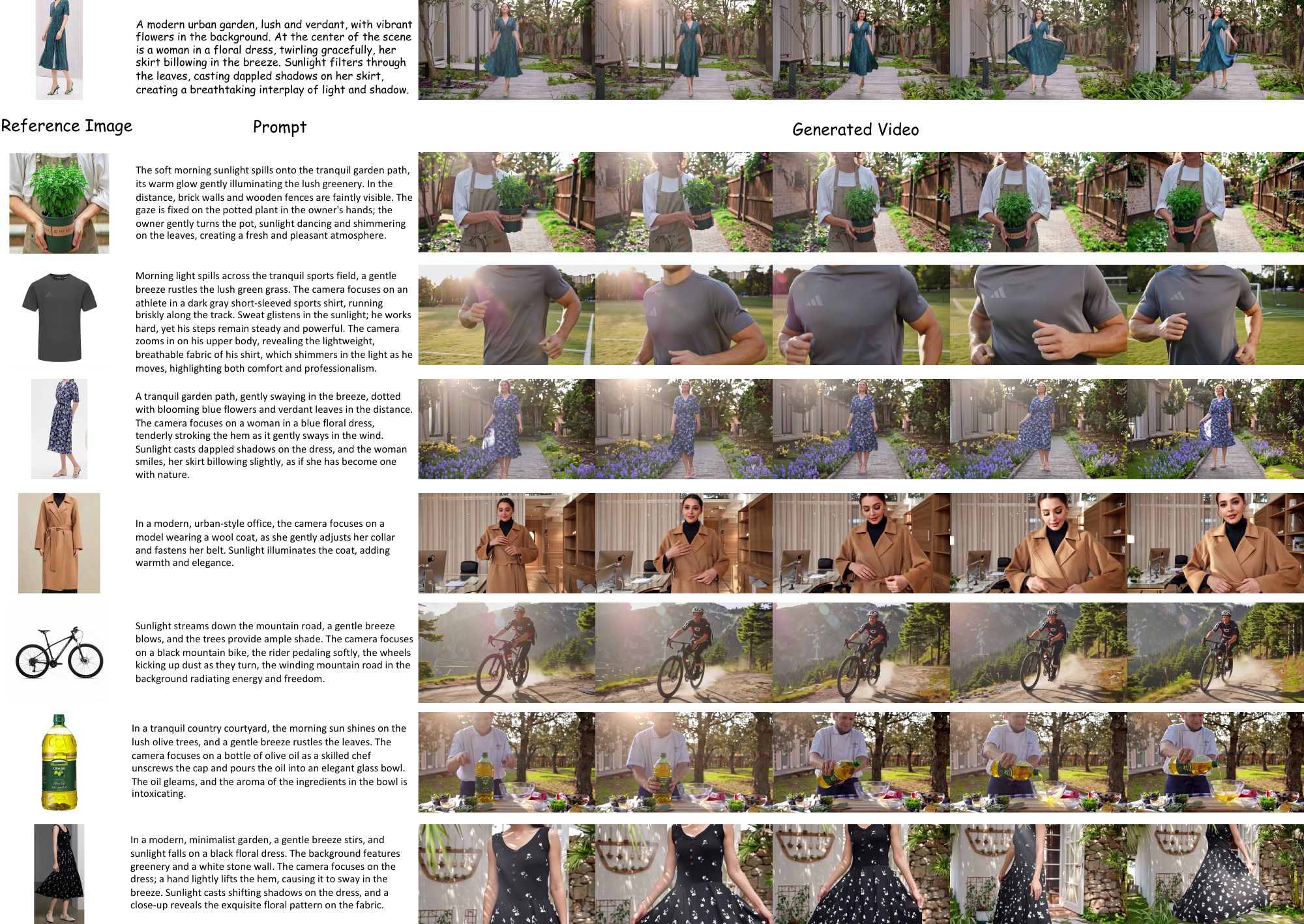}
\caption{\textbf{Reference Images to Video Results.} SkyReels-V3 can make advertising and product demonstration with one picture.}
\label{fig:mo2v4}
\end{figure*}

SkyReels-V3 can synthesizes temporally coherent video sequences conditioned on multiple visual references and a textual prompt. Given one to four reference images, which may correspond to characters, objects, or background scenes, the model generates videos that preserve identity attributes, spatial composition, and narrative continuity while following high-level semantic instructions.

\paragraph{Reference-Preserving Data Construction.}
For multi-reference image-guided video generation, the quality of the reference image-to-target video pairs is crucial. To this end, we introduce a dedicated data processing pipeline. 
Initially, we filter video data from a massive in house dataset, selecting clips that exhibit both high visual quality and significant dynamic motion.
Then, reference frames are selected from continuous video sequences using a cross-frame pairing strategy~\cite{fei2025skyreels,chen2025phantom}, ensuring temporal diversity while maintaining semantic consistency. 
Image editing models~\cite{labs2025flux,wu2025qwenimagetechnicalreport} are then applied to extract subject regions and perform background completion, together with semantic rewriting, to construct training pairs that avoid trivial frame copying and reduce copy-and-paste artifacts.
Furthermore, we have developed multiple filtering steps to remove distorted and inconsistent reference images generated by editing models.

\paragraph{Multi-Reference Conditioning.}
To effectively integrate heterogeneous reference inputs, SkyReels-V3 employs a unified multi-reference conditioning strategy that jointly encodes visual and textual information. 
For each reference image, we encode it using the video VAE and subsequently concatenate the resulting latent representation with the video latents.
By allowing up to four reference images, the model supports flexible scene composition and enables fine-grained control over subject appearance and background structure. 
This design facilitates complex multi-subject and multi-element video generation without requiring explicit manual composition.

\paragraph{Training Strategy.} 
We train the model using an image–video hybrid training scheme that jointly leverages large-scale image and video datasets. This approach enhances generalization by exposing the model to both static appearance cues and dynamic motion patterns. In addition, multi-resolution joint training is employed to improve robustness across different spatial scales and aspect ratios, enabling the model to natively support a wide range of output configurations.

\paragraph{Benchmark.}
\begin{table}[t]
\centering
\caption{Quantitative comparison of video generation models on reference consistency, instruction following, and visual quality. Higher is better.}
\label{tab:mo2v_generation_comparison}
\begin{tabular}{lccc}
\toprule
Model & Reference Consistency $\uparrow$ & Instruction Following $\uparrow$ & Visual Quality $\uparrow$ \\
\midrule
Vidu Q2     & 0.5961          & 27.84          & 0.7877 \\
Kling 1.6       & 0.6630          & 29.23          & 0.8034 \\
PixVerse V5   & 0.6542          & \textbf{29.34}          & 0.7976 \\
\textbf{SkyReels-V3} & \textbf{0.6698} & 27.22 & \textbf{0.8119} \\
\bottomrule
\end{tabular}
\end{table}
We construct a test set comprising 200 data pairs, with sources spanning scenarios such as film and television, e-commerce, and advertising. 
The types of reference images include those featuring characters, animals, objects, and background scenes.
The model's capabilities are evaluated from three primary perspectives: Reference Consistency, Instruction Following, and Visual Quality. 
Specifically, Reference Consistency encompasses aspects such as facial consistency, clothing consistency, object consistency, and background consistency. %
Visual Quality is assessed based on image quality, dynamic degree, aesthetic quality, and motion smoothness.
SkyReels-V3 has been benchmarked against leading contemporary models, with comparative results presented in Table~\ref{tab:mo2v_generation_comparison}.
The results demonstrate that the SkyReels-V3 model achieves state-of-the-art performance within the industry.
We also show qualitative results in Figure~\ref{fig:mo2v1}, Figure~\ref{fig:mo2v2}, Figure~\ref{fig:mo2v3}, and Figure~\ref{fig:mo2v4}.
These results indicate that the model demonstrates strong generalization capabilities across a wide range of scenarios.

\subsection{Video Extension}

\begin{figure*}[t]
\centering
\includegraphics[width=0.99\textwidth]{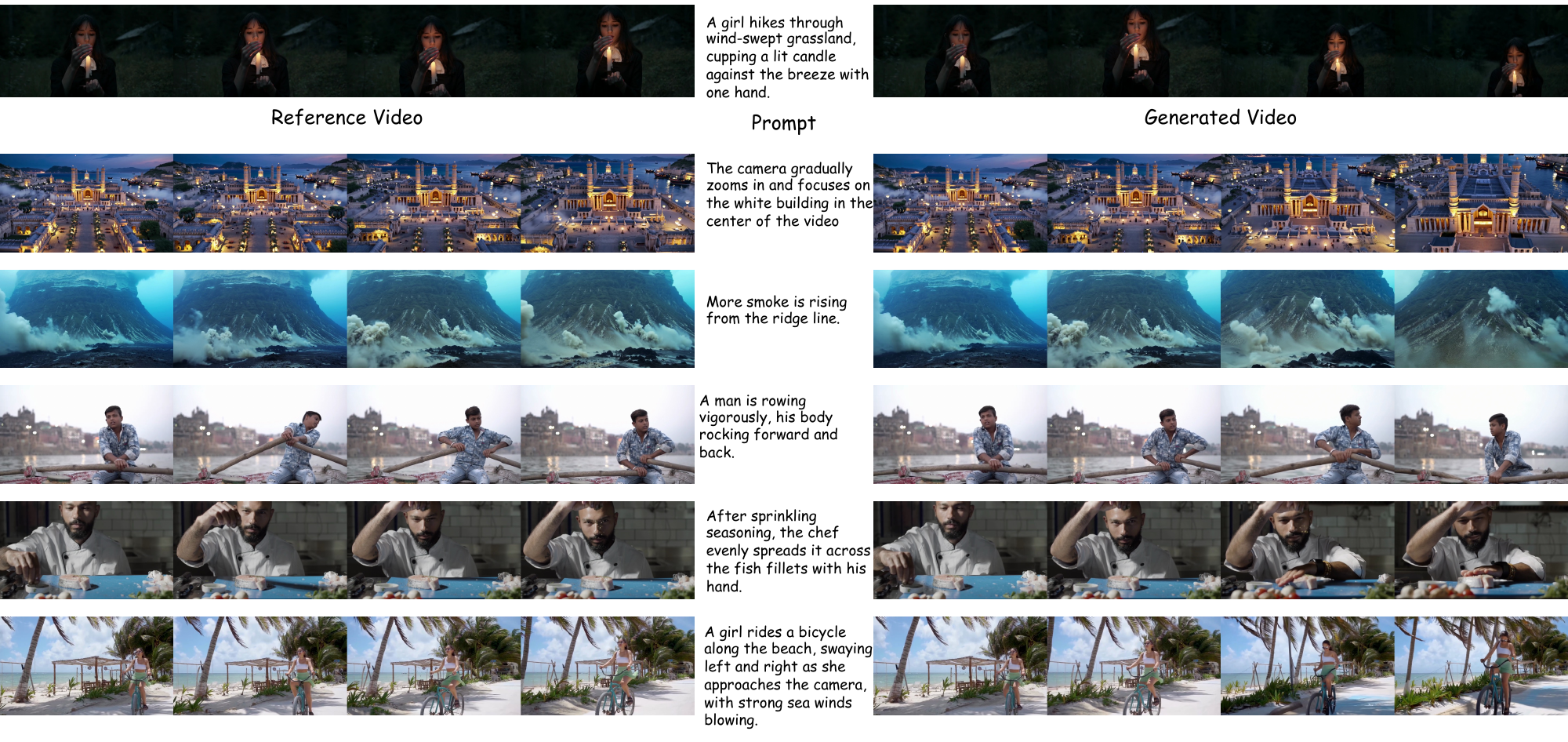}
\caption{\textbf{Single-shot Video Extension Results.} }
\label{fig:ve1}
\end{figure*}

The SkyReels-V3 Video Extension Model is designed to extend an input video segment into temporally coherent and semantically consistent subsequent content under textual guidance. 
Given an initial video clip, the model generates continuation segments that preserve motion dynamics, scene structure, and visual style, while maintaining narrative coherence across extended temporal horizons. 
It supports that: 
(i) \textbf{Dual extension modes.}
The model supports both single-shot video extension and shot switching video extension. For shot switching video extension, five predefined transition types are supported, and the mode can be selected either manually or through automatic detection. 
(ii) \textbf{High-quality visual synthesis.}
The system produces visually coherent extensions with stable composition, smooth motion, and seamless temporal continuity.
(iii) \textbf{Style-consistent generation.}
Visual style cues from the input video are explicitly preserved, enabling faithful continuation across realistic, cinematic, and domain-specific aesthetics.
(iv) \textbf{High-definition and flexible outputs. }
The model supports 720p video generation with adjustable extension durations ranging from 5 to 30 seconds for single-shot extension, as well as multiple aspect ratios (1:1, 3:4, 4:3, 16:9, and 9:16).

To achieve this, we introduce: (i) Shot switching detector. We have developed a shot switching detector to analyze long-form videos. It identifies whether shot transitions (cuts) are present and classifies their types. Currently supported transition types include \textbf{single shot}, \textbf{cut-in}, \textbf{cut-out}, \textbf{multi-angle}, \textbf{shot/reverse shot}, and \textbf{cut-away}. This detector enables the construction of effective training data. (ii) Unified multi-segment positional encoding and hierarchical training. A unified positional encoding scheme, combined with hybrid hierarchical data training, enables accurate motion modeling and smooth transitions across complex, multi-segment video extensions. (iii) Robust spatiotemporal modeling. The model effectively handles challenging scenarios, including rapid motion, multi-subject interactions, and abrupt scene changes, while enforcing physical plausibility and temporal consistency. 

\begin{figure*}[t]
\centering
\includegraphics[width=0.99\textwidth]{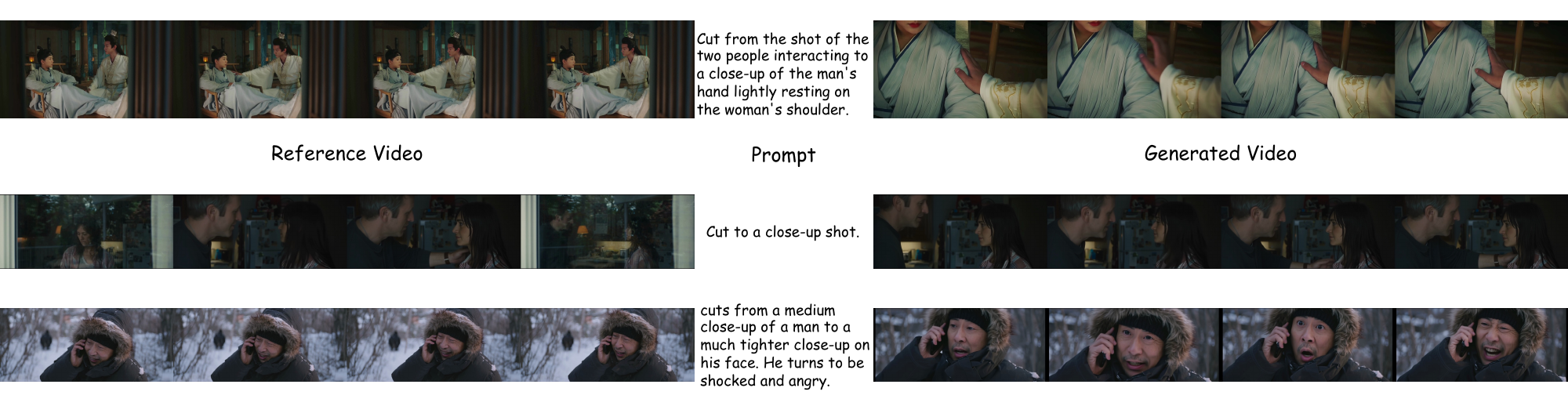}
\caption{\textbf{Shot-switching Video Extension (Cut In) Results.} }
\label{fig:cut_in}
\end{figure*}

\begin{figure*}[t]
\centering
\includegraphics[width=0.99\textwidth]{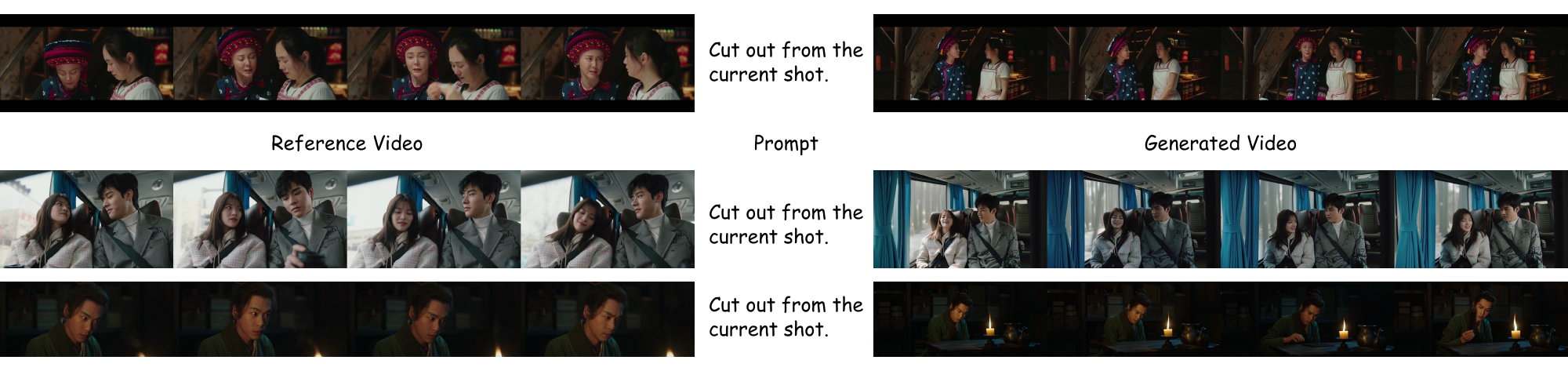}
\caption{\textbf{Shot-switching Video Extension (Cut Out) Results.} }
\label{fig:cut_out}
\end{figure*}

\begin{figure*}[ht]
\centering
\includegraphics[width=0.99\textwidth]{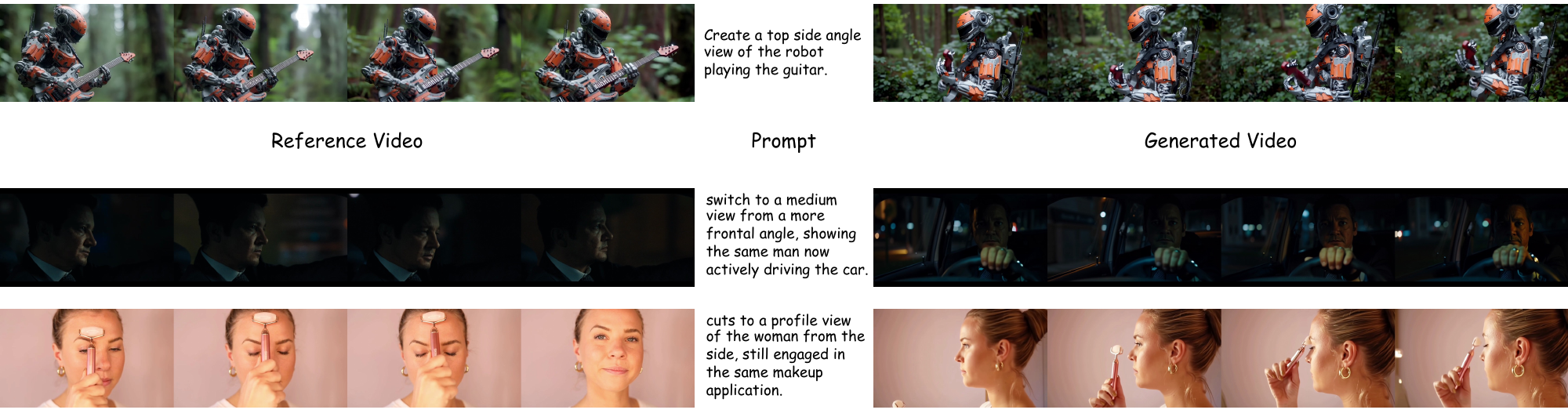}
\caption{\textbf{Shot-switching Video Extension (Multi Angle) Results.} }
\label{fig:multi_angle}
\end{figure*}

\begin{figure*}[t]
\centering
\includegraphics[width=0.99\textwidth]{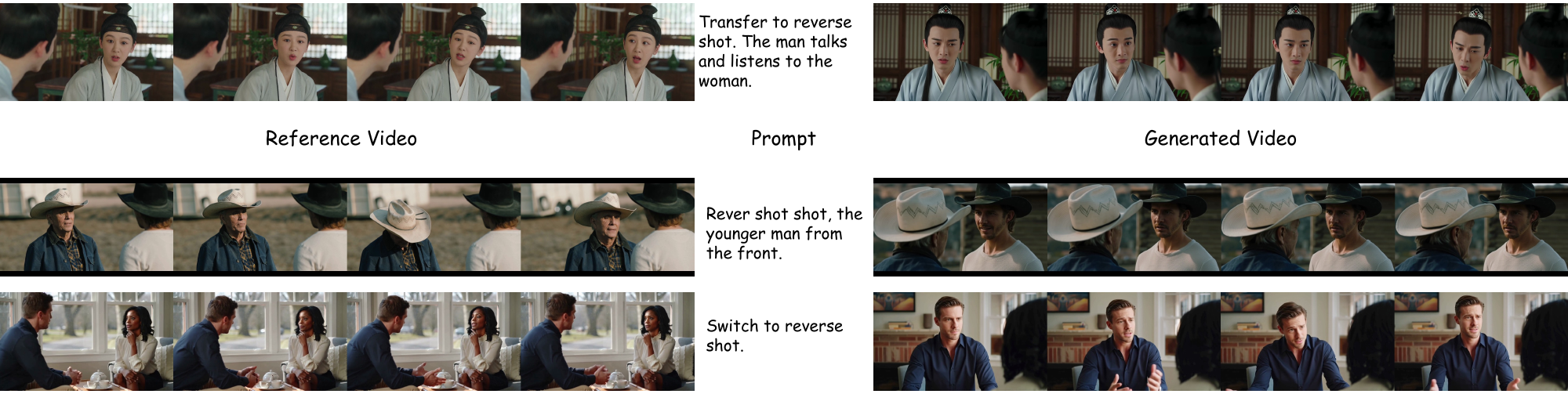}
\caption{\textbf{Shot-switching Video Extension (Shot/Reverse Shot) Results.} }
\label{fig:reverse}
\end{figure*}

\begin{figure*}[t]
\centering
\includegraphics[width=0.99\textwidth]{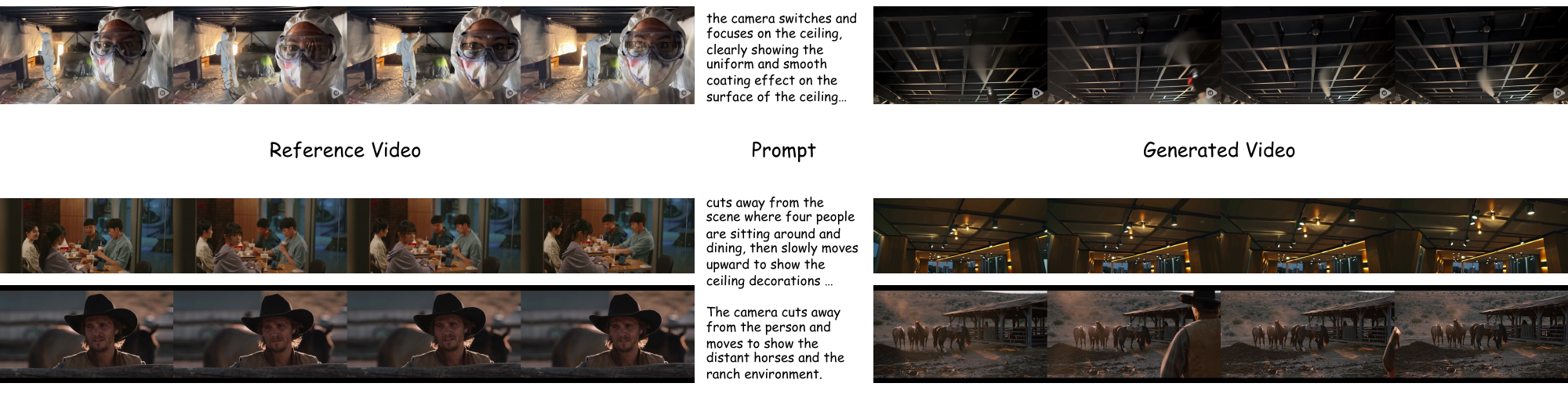}
\caption{\textbf{Shot-switching Video Extension (Cut Away) Results.} }
\label{fig:away}
\end{figure*}

Figure~\ref{fig:ve1} presents the results for single-shot extension, and Figure~\ref{fig:cut_in} to~\ref{fig:away} illustrate the extension results for cut-in, cut-out, multi-angle, shot/reserve shot, and cut-away.
The model generalizes well across diverse application scenarios, including cinematic content creation, short-form series production, game cutscenes, and long-form video enhancement, producing high-definition outputs with sharp visual details and natural motion dynamics.

\subsection{Talking Avatar}

Talking avatar model enables high-quality audio-conditioned video generation from a single portrait image and an input audio clip. 
The system is designed to produce temporally coherent, visually realistic videos with accurate audio–visual synchronization, supporting long-form generation and multi-character interactions.
It key improvements includes: 
(i) High-fidelity visual synthesis and precise lip synchronization.
The model can generate 720p videos at 24 fps, delivering smooth motion and fine-grained facial details. It supports multiple languages and speech types, ensuring that lip movements are closely aligned with phoneme-level audio dynamics, thereby enhancing realism and perceptual authenticity.
(ii) Multi-style character generalization. The framework is compatible with a wide range of visual styles, including photorealistic humans, cartoons, animals, and stylized characters. This flexibility enables broad applicability in virtual avatars, brand representation, and creative content generation.
(iii) Long-form coherent video generation. The model supports minute-level video synthesis in a single forward generation process, maintaining identity consistency, motion continuity, and expressive stability over extended durations. This makes it suitable for applications such as instructional videos, news narration, and long-form storytelling.
(iv) Multi-character scene support. The system is optimized for scenarios involving multiple avatars, allowing explicit role assignment and coordinated interactions. This capability facilitates the generation of dialogues, interviews, and other complex conversational scenes.
Note that in multi-person scenes, the mask must be used to specify which character is speaking.

Talking avatar model jointly analyzes audio signals, visual inputs, and textual cues to infer appropriate facial expressions, head movements, and camera dynamics, resulting in semantically and emotionally aligned video generative performances.
To achieve accurate lip synchronization, the model is trained using dedicated audio–visual alignment strategies with region masking that explicitly model the correspondence between speech units and facial motion. This design ensures robust performance across diverse languages, speaking styles, singing voices, and rapid speech patterns.
Furthermore, a key-frame-constrained generation framework is introduced to improve temporal stability in long videos. The model first establishes structurally important key content and then generates smooth transitions between key frames, ensuring consistent character appearance and natural motion flow throughout the sequence.
In internal evaluations against representative mainstream talking avatar models, talking avatar model demonstrates superior performance across multiple dimensions, including overall visual quality, lip synchronization accuracy, and expressive realism. The results indicate a clear advantage in producing stable, high-quality, and perceptually convincing talking avatar videos.

\begin{figure*}
\centering
\includegraphics[width=0.99\textwidth]{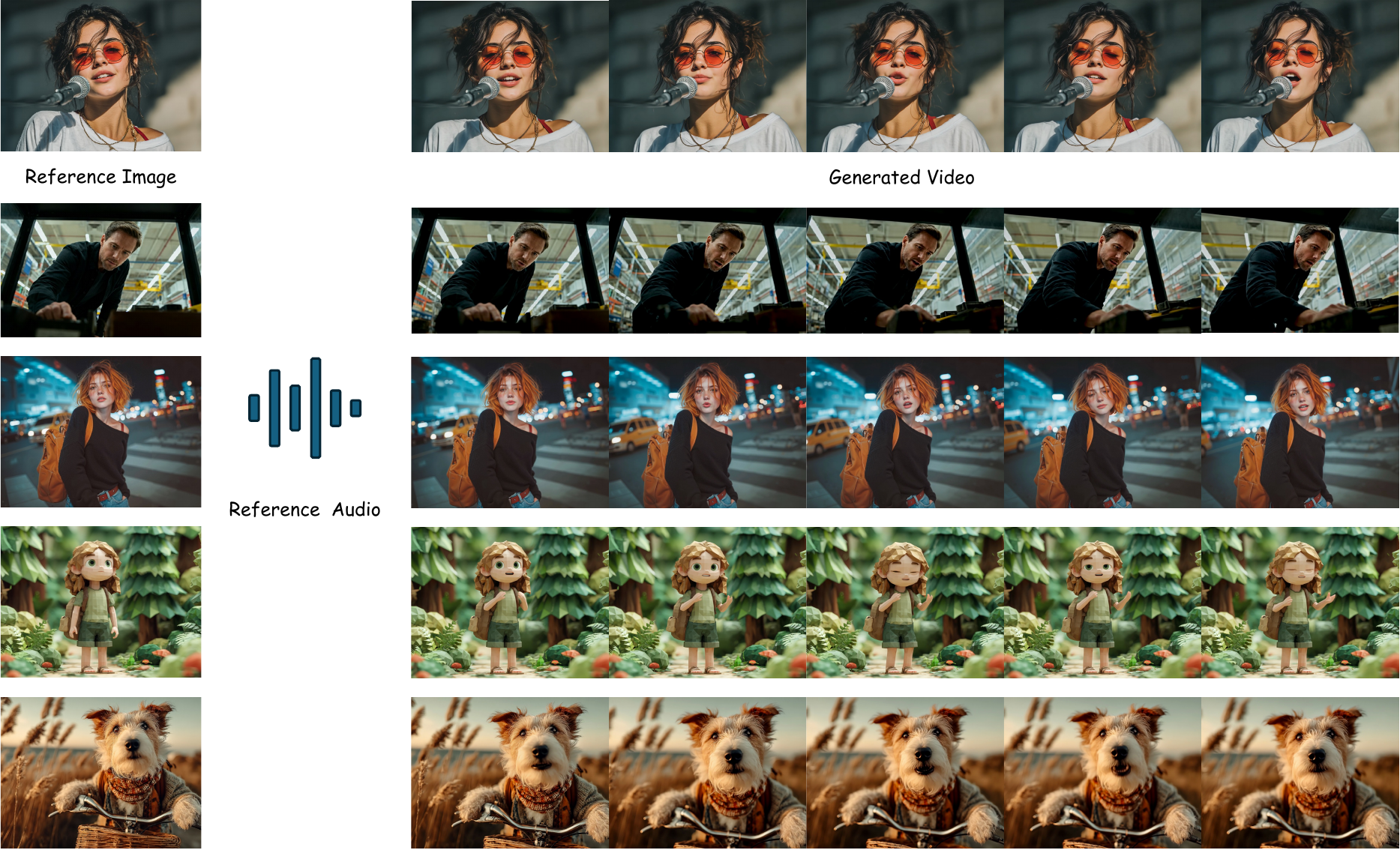}
\caption{\textbf{Generation results with multiple objectives and styles.} The results show the model's generalization across various non-human subjects as well as different styles.}
\end{figure*}

\begin{figure}
\centering
\includegraphics[width=0.99\textwidth]{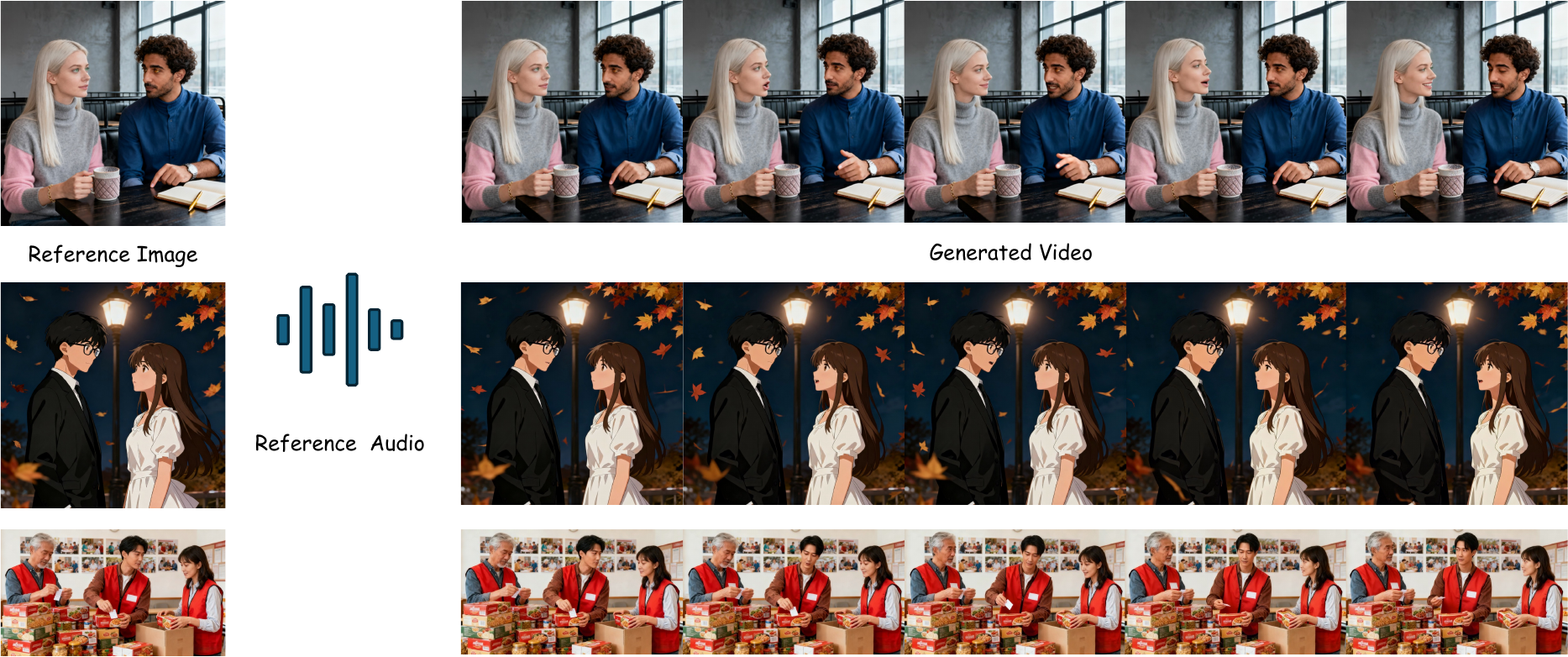}
\caption{\textbf{Multi-person results.} It presents a dialogue scenario, where characters correctly respond to conversational audio by switching between speaking and idle states. It showcase performance in multi-person scenes, with coordinated behavior for both speakers and listeners.}
\end{figure}

\begin{figure}
\centering
\includegraphics[width=0.99\textwidth]{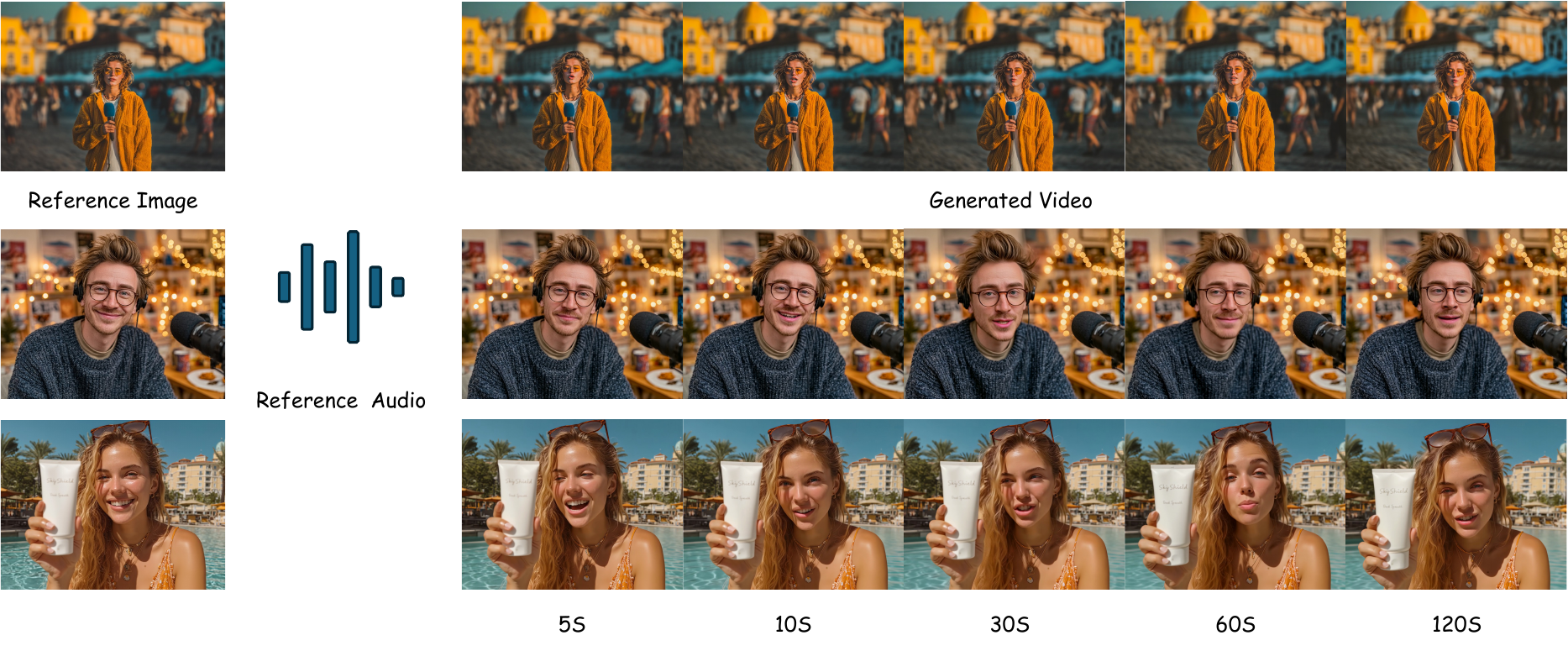}
\caption{\textbf{Minute-long video generation.} It show consistent visual effects with accurate audio alignment.}
\end{figure}

\begin{table}[t]
\centering
\caption{Quantitative comparison of talking avatar models on audio--visual synchronization, visual quality, and character consistency. Higher is better.}
\label{tab:avatar_comparison}
\begin{tabular}{lccc}
\toprule
Model & Audio--Visual Sync $\uparrow$ & Visual Quality $\uparrow$ & Character Consistency $\uparrow$ \\
\midrule
OmniHuman 1.5    & \textbf{8.25} & 4.60           & \textbf{0.81} \\
KlingAvatar          & 8.01          & 4.55           & 0.78          \\
HunyuanAvatar  & 6.72          & 4.50           & 0.74          \\
\textbf{SkyReels-V3} & 8.18          & \textbf{4.60} & 0.80          \\
\bottomrule
\end{tabular}
\end{table}

\section{Conclusion}

In this work, we presented SkyReels-V3, a unified multimodal video generation framework that integrates reference-based video synthesis, video extension, and audio-driven talking avatar generation within a single in-context learning paradigm. By jointly modeling visual, temporal, and auditory signals, SkyReels-V3 advances video generation from short, frame-level synthesis toward coherent, narrative-level content creation. Through innovations in multimodal conditioning, hybrid image–video training, hierarchical spatiotemporal modeling, and efficient token fusion strategies, the proposed system achieves strong subject consistency, high-fidelity motion generation, and robust instruction following across diverse tasks and aspect ratios. Extensive empirical evaluations demonstrate that SkyReels-V3 attains competitive performance on multiple benchmarks, rivaling leading closed-source models in various domain performance. Overall, SkyReels-V3 represents a significant step toward scalable, controllable, and general-purpose video generation systems, and provides a solid foundation for future research in multimodal generative modeling and cinematic-level video synthesis.

\bibliographystyle{plain}
\bibliography{main}

\clearpage
\appendix
\section*{Contributors and Acknowledgments}
Debang Li, ~ Zhengcong Fei, ~ Tuanhui Li, ~ Yikun Dou, ~ Zheng Chen, ~ Jiangping Yang, ~ Mingyuan Fan,
~ Jingtao Xu, ~  Jiahua Wang, ~ Baoxuan Gu, ~ Mingshan Chang, ~ Wenjing Cai, ~ Yuqiang Xie, ~ Binjie Mao, ~ Youqiang Zhang, ~ Nuo Pang, ~ Hao Zhang, ~ Yuzhe Jin, ~ Zhiheng Xu, ~ Dixuan Lin, ~ Guibin Chen, ~ Yahui Zhou

\end{document}